\documentclass{article}

\usepackage{PRIMEarxiv}

\usepackage[utf8]{inputenc} 
\usepackage[T1]{fontenc}    
\usepackage{hyperref}       
\usepackage{url}            
\usepackage{booktabs}       
\usepackage{amsfonts}       
\usepackage{nicefrac}       
\usepackage{microtype}      
\usepackage{lipsum}
\usepackage{fancyhdr}       
\usepackage{graphicx}       
\graphicspath{{media/}}     

\usepackage[utf8]{inputenc} 
\usepackage[T1]{fontenc}    
\usepackage{hyperref}       
\usepackage{url}            
\usepackage{booktabs}       
\usepackage{amsfonts}       
\usepackage{nicefrac}       
\usepackage{microtype}      
\usepackage{xcolor}         
\usepackage{mathtools} 
\usepackage{amsmath}
\usepackage{amssymb}
\usepackage{latexsym}
\usepackage[linesnumbered,ruled,vlined]{algorithm2e}
\usepackage{verbatim}
\usepackage{amsfonts, amsmath, amssymb, amsthm}
\usepackage{graphicx}
\makeatletter
\let\c@lofdepth\relax
\let\c@lotdepth\relax
\makeatother
\usepackage{subfigure}
\usepackage{array}
\usepackage{float}
\restylefloat{figure}
\usepackage{xcolor, graphicx}
\usepackage{multirow}
\usepackage{pdfpages}

\usepackage{draftwatermark}
\SetWatermarkText{Pre-print}
\SetWatermarkScale{0.7} 
\SetWatermarkColor[gray]{0.7}

\usepackage{tikz}
\usepackage{pgfplots}
\usepackage{tikz}
\usetikzlibrary{decorations.pathreplacing, decorations.markings}
\usetikzlibrary{fadings}
\usepgfplotslibrary{fillbetween,groupplots}
\usetikzlibrary{er,positioning,bayesnet}
\usetikzlibrary{shadows, arrows.meta, positioning, backgrounds}
\usepackage{filecontents}

\title{Deep-SITAR: A SITAR-Based Deep Learning Framework for Growth Curve Modeling via Autoencoders}

\author{ María Alejandra Hernández\\   
       Department of Mathematics\\
       University of the Basque Country\\
      \textit{ maahernandezve@gmail.com}
       \And
        Oscar Rodriguez  \\
        Neurologyca science and marketing\\
       \textit{osarodriguezme@unal.edu.co }
       \AND
        Dae-Jin Lee \\
        School of Science and Technology\\
       IE University\\
       \textit{daelee@faculty.ie.edu }
       }

\begin{document}

\maketitle

\begin{abstract}%
Several approaches have been developed to capture the complexity and nonlinearity of human growth. One widely used is the Super Imposition by Translation and Rotation (SITAR) model, which has become popular in studies of adolescent growth. SITAR is a shape-invariant mixed-effects model that represents the shared growth pattern of a population using a natural cubic spline mean curve while incorporating three subject-specific random effects --timing, size, and growth intensity-- to account for variations among individuals. In this work, we introduce a supervised deep learning framework based on an autoencoder architecture that integrates a deep neural network (neural network) with a B-spline model to estimate the SITAR model. In this approach, the encoder estimates the random effects for each individual, while the decoder performs a fitting based on B-splines similar to the classic SITAR model. We refer to this method as the Deep-SITAR model. This innovative approach enables the prediction of the random effects of new individuals entering a population without requiring a full model re-estimation. As a result, Deep-SITAR offers a powerful approach to predicting growth trajectories, combining the flexibility and efficiency of deep learning with the interpretability of traditional mixed-effects models.

\end{abstract}

\section{Introduction}

Modeling growth curves is essential to understand and analyze human growth patterns in various areas, including medicine, psychology, and sports \cite{cole2011chart,psic_levesque2008predicting,malina2008growth}. Its purposes include monitoring changes across time, diagnosing growth disorders, improving the physical development of young athletes, and preventing injuries during growth phases.  Understanding the overall growth process has led to the development of multiple mathematical and statistical models \cite{davidian2003nonlinear, hauspie2004methods, hernandez2023derivative}, which typically rely on longitudinal data that consist of repeated measurements of each individual over time. Therefore, modeling growth curves requires statistical methods that retain the characteristics of the data and accurately capture the complexity of human growth.

One of the widely used methodologies for analyzing longitudinal data is the mixed-effects model for its ability to account for both population-level trends and individual-specific variations \cite{lindstrom1990nonlinear, pinheiro2006mixed}. Acknowledging the nonlinear nature of the human growth processes, nonlinear mixed-effects models have become essential tools for addressing this complexity. They offer flexibility, enhance interpretability, and account for the correlation inherent in longitudinal data \cite{grimm2011nonlinear,elhakeem2022using,cole2010sitar}. 

In parallel, the growing interest in machine learning methods has opened new possibilities for analyzing growth curves. These approaches are particularly attractive due to their strong prediction capabilities and flexibility in handling complex, high-dimensional data. These methods offer more efficient and accurate solutions for handling new data and often improve computational performance \cite{roush2006comparison,panagou2007application,ahmad2009poultry}.  Deep learning models based on neural networks allow non-linear modeling without requiring extensive knowledge about the relationships between variables, such as predefined functional forms or assumptions of linearity \cite{samek2021explaining}. However, classical neural networks assume that observations are independent \cite{cascarano2023machine,pinkus1999approximation}, which introduces biases when applied to correlated longitudinal data and leads to problems estimating individual-level effects or accurately capturing temporal dependencies.

Diverse models have been proposed within the neural network framework to address the challenge of correlations in variables estimated using neural networks and improve their applicability to repeated measures and hierarchical data. For example, in \cite{tandon2006neural}, authors introduced a mixed-effects neural network (MENN) that accounts for correlated observations that consider correlated observations. They focus on normally distributed responses and networks with no hidden layers; in \cite{tran2017random}, authors incorporate neural networks into the mixed model framework through a flexible generalized linear mixed model and handle estimation in a Bayesian context using variational approximation methods. Additionally, \cite{mandel2023neural} proposed a generalized neural network mixed model (GNMM) incorporating feed-forward neural networks into the mixed model framework and \cite{ramchandran2021longitudinal} proposed the longitudinal variational autoencoder (L-VAE), which incorporates a multi-output additive Gaussian process to model the correlation structure between samples based on auxiliary information. Despite these advances, applying neural networks to repeated measures and longitudinal data remains challenging. These challenges emphasize the need for further exploration to develop models that effectively combine the predictive power of neural networks with the rigorous statistical properties required for longitudinal data analysis \cite{tran2017random,maity2013subject}.

In this work, we focus on integrating the mixed-effects modeling approach with a neural network framework to analyze growth curves. Specifically, we study the Superposition by Translation and Rotation (SITAR) model, introduced by \cite{cole2010sitar}, which has gained popularity for modeling individual growth trajectories. SITAR is a nonlinear, shape-invariant mixed-effects model that represents individual growth patterns relative to a common underlying curve. It uses a mean natural cubic spline to capture the overall growth trend and includes three individual-specific random effects:  a shift in timing, a scaling of growth magnitude, and a shift in size.  The goal is to offer alternatives to the SITAR model because SITAR's disadvantage for growth curve analysis is that it lacks the ability to predict the growth trajectories of new individuals entering a population.

We introduce Deep-SITAR, an autoencoder-based model that extends the classical SITAR model. In this approach, we replace the natural cubic spline with a B-spline basis, and the random effects are estimated using a neural network to fit the growth curves. This approach combines the flexibility of B-splines with the improved predictive capabilities of neural networks while preserving the ability of mixed-effect models to capture individual variability. We apply Deep-SITAR to growth curve modeling and compare its performance with the classical SITAR, using simulated data based on the Berkeley dataset available in \texttt{R}. Deep-SITAR is implemented in \texttt{Python} using the \texttt{TensorFlow} library. In contrast, the classical SITAR is implemented in statistical software \texttt{R} \cite{Rmanual} using the \texttt{sitar} library, allowing a straightforward comparison of their computational and statistical properties.

The structure of this work is as follows: Section \ref{sec:background} provides background on mixed-effects models, and Section \ref{sec:sitar_model} describes the classic SITAR model. Section \ref{sec:neural network} introduces the neural networks framework and explains the autoencoder architecture. Section \ref{sec:Deep_sitar} presents the implementation of Deep-SITAR, which integrates the mixed-effects model with the neural network framework. Section \ref{sec:simulation} discusses an experiment with simulated data to assess and compare the model performance. Finally, in Section \ref{sec:conclusion}, we discuss the results and propose future directions for research.

\section{Mixed-effect models} \label{sec:background}

The mixed-effects model extends fixed effects models by incorporating random effects, allowing for the analysis of both between-subject and within-subject variation in longitudinal data. These models consist of two components: (i) fixed effects, representing population-level parameters that remain constant among individuals, and (ii) random effects, which account for individual variations and are assumed to follow a known probability distribution.

\begin{figure}[ht!]
\centering
\includegraphics[width=\linewidth, page=1]{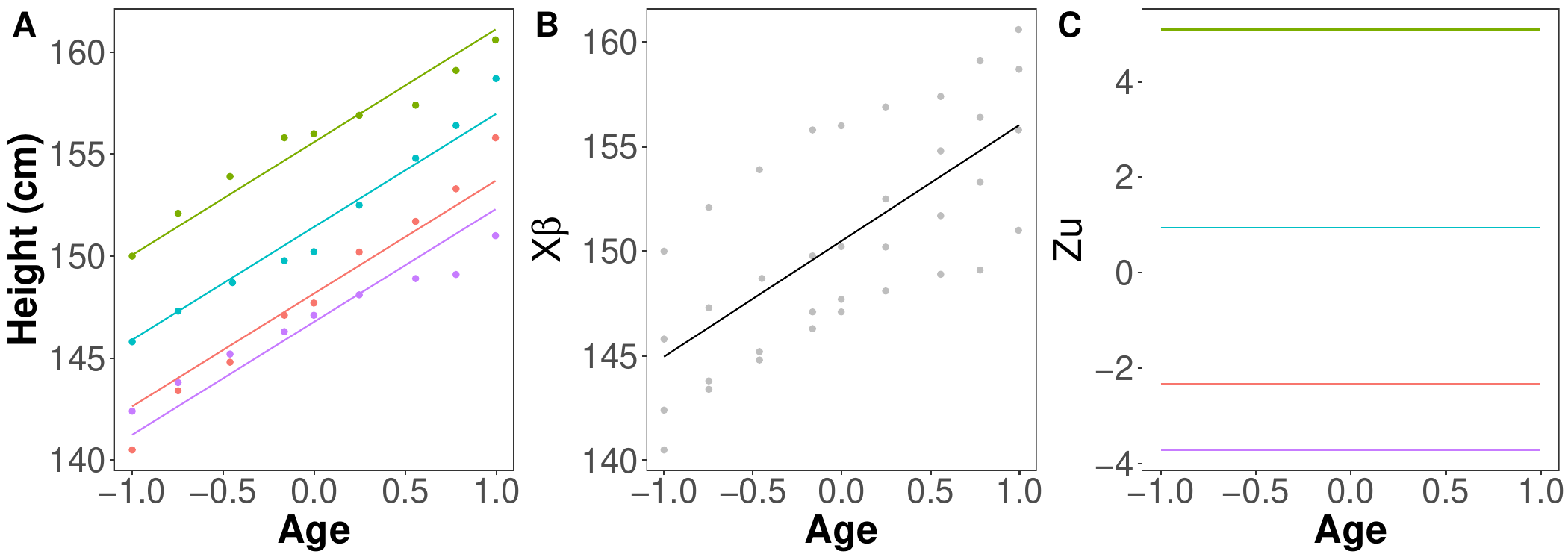}
\caption{Example of growth curves with four fitted individuals using the LME. \textbf{A:} Fitted curves. \textbf{B:} Population curve. \textbf{C:} Individual deviation curves from the overall trend.}
\label{fig:lme_m}
\end{figure}

A common form of mixed-effects models is the linear mixed model (LME), which extends the classical linear model by incorporating fixed and random effects, as shown in Figure \ref{fig:lme_m}. Consider a set of $N$ individuals measured at the same time points, where $y_{ij}$ is the observation of the $i$-th individual taken at time $t_{ij}$. The linear mixed model for the $i$-th individual is expressed as:
\begin{equation}
    y_{ij}= (\beta_{0}+u_{0i})+(\beta_{1}+u_{1i})t_{ij}+\varepsilon_{ij}, \quad i = 1, ..., N, \quad j=1, ..., n \label{firsteq}
\end{equation}

In this case, each individual parameter is additively decomposed into a fixed effect $\beta_k$ and a random effect $u_{ki}$ with $k=0,1$. We rewrite in matrix form for the $i$-th individual the expression in Equation \eqref{firsteq} as follows: 
\begin{equation}\label{eq:LME}
     \boldsymbol{y}_i =\boldsymbol{X}_i\boldsymbol{\beta}+\boldsymbol{Z}_i\boldsymbol{u}_i+\boldsymbol{\varepsilon}_i, \quad\boldsymbol{\varepsilon}_i \sim \mathcal{N}(\boldsymbol{0}, \sigma^2 \boldsymbol{I}), \quad \boldsymbol{u}_i\sim \mathcal{N}(\boldsymbol{0}, \boldsymbol{\Lambda})
\end{equation}

where $\boldsymbol{\beta}=(\beta_0, \beta_1)^\top$ and $\boldsymbol{u}_i=(u_{0i}, u_{1i})^\top \sim \mathcal{N}(\boldsymbol{0},\boldsymbol{\Lambda})$ are the fixed and random effect vectors, respectively. The errors $\boldsymbol{\varepsilon}_i$ for $i=1,..., N$, follow a normal distribution with mean zero and variance $\sigma^2 \boldsymbol{I}$, and the random effects and errors are independent.

The model fitting consists of three key components: estimating fixed effects, predicting random effects, and estimating variance parameters.  The LME model in Equation \eqref{eq:LME} is fitted by maximizing the  log-likelihood that is based on the joint density of $(\boldsymbol{y},\boldsymbol{u})$, given by \cite{pawitan2001all} :
\begin{align}\label{eq:lme_l}
    \ell(\boldsymbol{\beta},  \sigma, \boldsymbol{\Lambda},\boldsymbol{u}|\boldsymbol{y}) &\propto -\frac{1}{2 \sigma^2}\|\boldsymbol{y}-\boldsymbol{X}\boldsymbol{\beta}-\boldsymbol{Z}\boldsymbol{u}\|^2_2- \frac{1}{2}\boldsymbol{u}^\top \boldsymbol{\Lambda}^{-1}\boldsymbol{u}
\end{align}

However, LME models do not always provide the best description of the data. In certain cases, such as growth model analysis, nonlinear and semiparametric approaches are more suitable \cite{lindstrom1990nonlinear, pinheiro2006mixed, ke2001semiparametric}. In this work, we study the SITAR model to analyze longitudinal growth curves. While traditional models, such as the Preece-Baines \cite{preece1978new} and Gompertz models \cite{winsor1932gompertz}, describe growth patterns using predefined functional forms, these models may not fully capture individual variations. In contrast, SITAR offers a more flexible approach. It summarizes growth trajectories with biologically meaningful parameters, capturing individual differences and adapting to various patterns without assuming a predefined shape. The next section provides a detailed explanation of the SITAR model.

\subsection{SITAR} \label{sec:sitar_model}

The SITAR model \cite{cole2010sitar} is a non-linear shape-invariant mixed-effects model \cite{lindstrom1995self, beath2007infant} that assumes the population shares a common growth curve shape, such that shifting and scaling this curve is made to match the individual growth patterns. The SITAR model is represented as follows:

 \begin{align}\label{eq:sitar_model}
     y_{ij}&= a_i + S\left(\frac{t_{ij}- b_i}{e^{-c_i}} \right)+\epsilon_{ij},  \qquad i=1, ..., N, \quad j=1,..., n\
 \end{align}

where $y_{ij}$ is the height of the $i$-th individual at the $j$-th measurement at time $t_{ij}$, $S(\cdot)$ is a natural cubic spline function over time, providing the nonlinearity and base shape of the height curves, $a_i= a_0+a_{1i}, \ b_i= b_0+b_{1i}$ and $c_i=c_0+c_{1i}$ are the individual parameters, where $a_0, b_0$ and $c_0$ are fixed effects and $a_{1i}, b_{1i}$ and $c_{1i}$ are normally distributed random effects with mean $\boldsymbol{0}$ and variance-covariance matrix $\boldsymbol{\Lambda}$.  The error $\epsilon_{ij}$ is assumed to be independent of the random effects and normally distributed with mean $0$ and variance $\sigma^2$.  

\begin{figure}[ht!]
\centering
\includegraphics[width=1\linewidth, page=1]{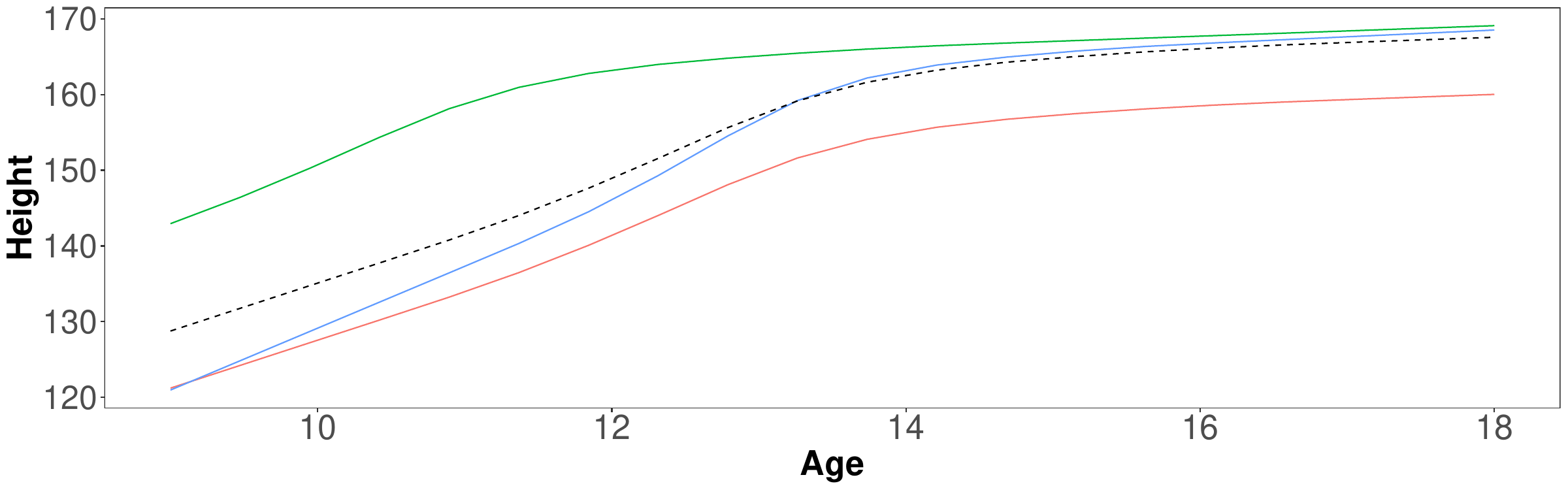}
\caption{SITAR features. The black dashed line represents the mean growth curve, while the colored curves illustrate the effects of the random effects. The red line indicates the height shift corresponding to $a_{1i}$, the green line corresponds to the age shift corresponding to $b_{1i}$, and the blue line represents the shrinking–stretching of the age scale corresponding to $c_{1i}$. }
\label{fig:sitar_re}
\end{figure}

The random effects $\boldsymbol{u}_i=(a_{1i},b_{1i},c_{1i})$, represent distinct transformations of the mean curve, where $a_{1i}$ represents a vertical shift, $b_{1i}$ indicates a horizontal shift, and $b_{1i}$ corresponds to the scaling (shrinking or stretching along the age axis) and rotation of the curve, as depicted in Figure \ref{fig:sitar_re}.  In addition, the SITAR model is widely used to study height growth patterns during puberty, and its random effects have clear biological interpretations \cite{cole2010sitar}. Specifically, the random effect $a_{1i}$ captures the difference in size relative to the mean height, $b_{1i}$ captures the timing difference of the pubertal growth spurt across individuals, and $c_{1i}$ captures the growth velocity difference across individuals. Consequently, these three random effects are called size, tempo, and velocity, respectively.

Additionally, a quantity called the effective degrees of freedom $(df)$ is used to quantify the complexity of the spline model. The $df$  represents the equivalent number of parameters effectively employed to fit the model. In practice, the model is estimated using Linstrom and Bates approximation  \cite{lindstrom1990nonlinear} of the maximum likelihood that is implemented in the \texttt{sitar} library --based on \texttt{nlme} and \texttt{ns} functions-- in \texttt{R}.

In the following section, we aim to connect mixed-effects models with deep neural networks. To this end, we introduce classical neural networks applied to the prediction of random effects $\boldsymbol{u}_i$ from measurements  $y_i$  for each individual, where $i=1,..., N$.

\section{Mixed-effects with deep neural networks} \label{sec:neural network}
A neural network is a function composed of multiple nested functions that map an input to an output by successive layers of transformations. The objective is to adjust the neural network parameters $\boldsymbol{\theta}$ --weights and biases-- so that the neural network's output closely approximates the true value as well as possible. \cite{goodfellow2016deep,samek2021explaining, rodriguez2023multimodal}.

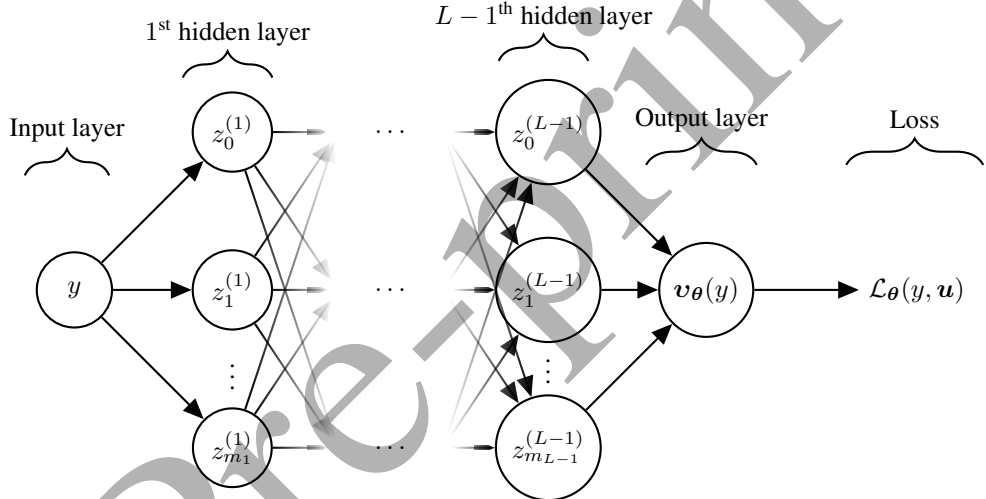
\begin{figure}[H]
\centering
\begin{tikzpicture}[thick, scale=0.7]
    \tikzstyle{unit0}=[draw,shape=circle,minimum size=1.cm]
	\tikzstyle{unit}=[draw,shape=circle,minimum size=1.cm]
	\tikzstyle{unit2}=[draw,shape=circle,minimum size=1.2cm]
	\tikzstyle{hidden}=[draw,shape=circle,minimum size=1.cm]
	\tikzstyle{FP}=[draw,shape=rectangle,minimum size=1.2cm]

	\node[unit](x0) at (0,1){$y$};

	\node[hidden](h10) at (3,4){$z_0^{(1)}$};
	\node[hidden](h11) at (3,1){$z_1^{(1)}$};
	\node at (3,-0.5){\vdots};
	\node[hidden](h1m) at (3,-2){$z_{m_1}^{(1)}$};

	\node(h22) at (5,-2){};
	\node(h21) at (5,1){};
	\node(h20) at (5,4){};
	
	\node(d3) at (6,-2){$\ldots$};
	\node(d2) at (6,1){$\ldots$};
	\node(d1) at (6,4){$\ldots$};

	\node(hL12) at (7,-2){};
	\node(hL11) at (7,1){};
	\node(hL10) at (7,4){};
	
	\node[hidden](hL0) at (9,4){$z_0^{(L-1)}$};
	\node[hidden](hL1) at (9,1){$z_1^{(L-1)}$};
	\node at (9,-0.4){\vdots};
	\node[hidden](hLm) at (9,-2){$z_{m_{L-1}}^{(L-1)}$};

	\node[unit](y1) at (12,1){$\boldsymbol{\upsilon}_{\boldsymbol{\theta}}(y)$};
	
	\node[](yend) at (16,1){$\mathcal{L}_{\boldsymbol{\theta}}(y,\boldsymbol{u})$};
	
    \draw[->] (x0) -- (h10);
	\draw[->] (x0) -- (h11);
	\draw[->] (x0) -- (h1m);

	\draw[->] (hL0) -- (y1);
	
	\draw[->] (hL1) -- (y1);
	
	\draw[->] (hLm) -- (y1);
	
    \draw[->,path fading=east] (h10) -- (h20);
	\draw[->,path fading=east] (h10) -- (h21);
	\draw[->,path fading=east] (h10) -- (h22);
	
	\draw[->,path fading=east] (h11) -- (h20);
	\draw[->,path fading=east] (h11) -- (h21);
	\draw[->,path fading=east] (h11) -- (h22);
	
	\draw[->,path fading=east] (h1m) -- (h20);
	\draw[->,path fading=east] (h1m) -- (h21);
	\draw[->,path fading=east] (h1m) -- (h22);
	
	\draw[->,path fading=west] (hL10) -- (hL0);
	\draw[->,path fading=west] (hL11) -- (hL0);
	\draw[->,path fading=west] (hL12) -- (hL0);
	
	\draw[->,path fading=west] (hL10) -- (hL1);
	\draw[->,path fading=west] (hL11) -- (hL1);
	\draw[->,path fading=west] (hL12) -- (hL1);
	
	\draw[->,path fading=west] (hL10) -- (hLm);
	\draw[->,path fading=west] (hL11) -- (hLm);
	\draw[->,path fading=west] (hL12) -- (hLm);
	
	\draw[->] (y1) -- (yend);
	
	\draw [decorate,decoration={brace,amplitude=10pt},xshift=-4pt,yshift=5pt] (-0.8,3) -- (0.8,3) node [black,midway,yshift=+0.6cm]{Input layer};
	\draw [decorate,decoration={brace,amplitude=10pt},xshift=-4pt,yshift=1.0pt] (2.2,5) -- (3.9,5) node [black,midway,yshift=+0.6cm]{$1^{\text{st}}$ hidden layer };
	\draw [decorate,decoration={brace,amplitude=10pt},xshift=-4pt,yshift=10pt] (8.2,5) -- (9.9,5) node [black,midway,yshift=+0.6cm]{$L-1^{\text{th}}$ hidden layer };
	\draw [decorate,decoration={brace,amplitude=10pt},xshift=-4pt,yshift=5pt] (11,3.2) -- (13.1,3.2) node [black,midway,yshift=+0.6cm]{Output layer};
	\draw [decorate,decoration={brace,amplitude=10pt},xshift=-4pt,yshift=5pt] (14.8,3.2) -- (17.4,3.2) node [black,midway,yshift=+0.6cm]{Loss};
\end{tikzpicture}
\caption{Fully connected neural network scheme.}
\label{FCNN}
\end{figure}

In our proposed approach, we aim to map the input measurements $\boldsymbol{y}_i$ to the continuous unobserved random effects $\boldsymbol{u}_i$ for $i = 1, ..., N$, fitting a classical neural network to predict random effects based on the given measurements.  We consider a classical neural network as a function \(\boldsymbol{\upsilon}_{\boldsymbol{\theta}}(.)\) with parameters \(\boldsymbol{\theta}\) that fits \(\boldsymbol{u}_{i} = \boldsymbol{\upsilon}_{\boldsymbol{\theta}}(\boldsymbol{y}_i) + \boldsymbol{\varepsilon}_i\), where \(\boldsymbol{\varepsilon}\) represents the estimation error between predictions and observations. Figure \ref{FCNN} shows a fully connected neural network scheme with $L$ layer, and mathematically a neural network $\boldsymbol{\upsilon}_{\boldsymbol{\theta}}$ takes the form:

\begin{equation}
    \boldsymbol{\upsilon}_{\boldsymbol{\theta}}(y) = (\mathfrak{L}^{(L)}_{\theta_{L}}\circ\mathfrak{L}^{(L-1)}_{\theta_{L-1}}\circ\cdots\circ\mathfrak{L}^{(1)}_{\theta_{1}})(y),
\end{equation}
each \(k\)-th layer $\mathfrak{L}^{(k)}_{\theta_{k}}(y)$ contains the parameters $\boldsymbol{\theta}_{k}$ and \(m_k\) nodes  $\{ z_0^{(k)}, ... ,  z_{m_k}^{(k)}\}$. For the first layer, the nodes are given by :
\begin{equation}
z_l^{(1)}=  \mathfrak{F}_{1}\left(w_l^{(1)}y+d_l^{(1)}\right),\quad l=1, ..., m_1,    
\end{equation}
where $w_l^{(1)}$ is a weight, $d_l^{(1)}$ is a bias and $\mathfrak{F}(\cdot)$ is known as the \textit{activation function}. Therefore, for the $k$-th hidden layer $(2\leq k \leq L-1)$, each node is defined as:

\begin{equation}
z_l^{(k)}= \mathfrak{F}_{k}\left(\sum_l^{m_{k-1}} w_l^{(k)}z_l^{(k-1)}+d_l^{(k)}\right),  \quad l=1, ..., m_k,  
\end{equation}
with $\boldsymbol{\theta}_k = \{w_l^{(k)},d_l^{(k)}\}_{l=1}^{m_{k-1}}$.

In deep learning, the objective function to be minimized is called the \textit{loss function}. In this case, to estimate $\boldsymbol{\theta}$, training aims to minimize a combination of loss and model complexity. A commonly used loss function is the Mean Squared Error (MSE). To quantify complexity, we incorporate the regularization term from Equation \eqref{eq:LME}. Thus, the objective function is defined as follows \cite{hu2023improvement,wang2022comprehensive}:

\begin{equation}
    \mathcal{L}_{\boldsymbol{\theta}}(\boldsymbol{y}_i,\boldsymbol{u}) := \frac{1}{N}\sum_{i=1}^N\left(\|\boldsymbol{\upsilon}_{\boldsymbol{\theta}}(\boldsymbol{y}_i)-\boldsymbol{u}_i\|_2^2 + \boldsymbol{\upsilon}_{\boldsymbol{\theta}}(\boldsymbol{y}_i)^\top\boldsymbol{\Lambda}_{\boldsymbol{ \upsilon}}^{-1}\boldsymbol{\upsilon}_{\boldsymbol{\theta}}(\boldsymbol{y}_i)\right) \label{MSE1}
\end{equation}
where $\boldsymbol{\Lambda}_{\boldsymbol{ \upsilon}}$ is the estimated variance-covariance matrix of the $\boldsymbol{\upsilon}_{\boldsymbol{\theta}}(y)$.

The process of incorporating mixed effects into the neural network framework begins when the measurement vector $\boldsymbol{y}_i$ for the  $i$-th  individual enters the neural network through the first hidden layer. It undergoes a linear transformation followed by a nonlinear mapping through an activation function. The transformed values, represented by the $m_1$ nodes $z^{(1)}$, are then passed through successive layers, each applying the same transformation until the expected random effects output $\boldsymbol{u}_i$ is obtained.

As part of our objective, we aim to estimate the random effects simultaneously with the fit. However, this is not possible using a classical neural network. Therefore, in the next section, we briefly introduce autoencoders and present our proposed autoencoder approach, which combines B-splines and neural networks.

\subsection{Autoencoder}
An autoencoder is the assembly of two architectures. The first of these is called an \textit{encoder} ($\boldsymbol{\upsilon}(y)$), where the output ($\boldsymbol{u}$) of this is the input of another architecture called a \textit{decoder} ($\mathcal{S}(t,\boldsymbol{u})$), and which, in turn, the output ($y$) of the decoder is the input of the encoder. In this way, when assembled, the output of the autoencoder is an estimation of its input ($\mathcal{DS}(t,y)=\mathcal{S}\circ\boldsymbol{\upsilon}(t,y)=\widehat{y}$). In addition, it has a latent variable ($\boldsymbol{u}$), which is the output and input of the assembly ($\mathcal{S}(t,\boldsymbol{\upsilon}(y)=\widehat{y})=\widehat{y}$). This style of architecture has a significant advantage over traditional models because the latent variable is estimated during the adjustment. Usually, no data are needed since it does not directly affect the loss function, as we show below:

\begin{equation}
    \mathcal{L}(t,y) = \|\mathcal{DS}(t,y)-\widehat{y}\|^2_2 = \|\mathcal{S}(t,\widehat{y})-\widehat{y}\|^2_2 = \|\mathcal{S}\circ\boldsymbol{\upsilon}(t,y)-\widehat{y}\|^2_2. \label{nnre}
\end{equation}

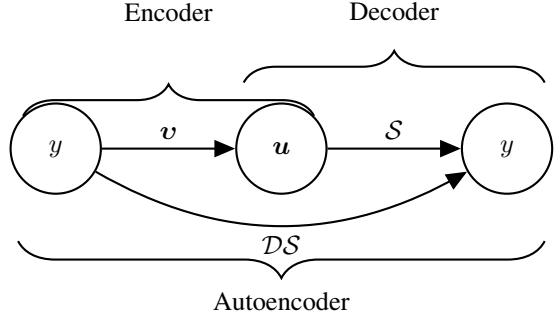
\begin{figure}[h]
    \centering
    \begin{tikzpicture}[thick, scale=1, every node/.style={transform shape}]
        \tikzstyle{unit}=[draw,shape=circle,minimum size=1.2cm]
        \tikzstyle{hidden}=[draw,shape=circle,minimum size=1.2cm]

        \node[unit] (x1) at (0, 0) {$y$};
        \node[hidden] (y) at (3, 0) {$\boldsymbol{u}$};
        \node[unit] (x2) at (6, 0) {$y$};

        \draw[->] (x1) -- node[above] {$\boldsymbol{\upsilon}$} (y);
        \draw[->] (y) -- node[above] {$\mathcal{S}$} (x2);
        \draw[->, bend right=30] (x1) to node[below] {$\mathcal{DS}$} (x2);
        
        \draw [decorate, decoration={brace, amplitude=12pt}, yshift=45pt] (x1.north west) -- (y.north east) node [midway, above=32pt] {Encoder};
        \draw [decorate, decoration={brace, amplitude=12pt}, yshift=25pt] (2.5,0) -- (6.5,0)  node [midway, above=20pt] {Decoder};
        \draw [decorate, decoration={brace, amplitude=14pt, mirror}, yshift=-35pt] (-0.5,0) -- (6.5,0) node [midway,yshift=-.8cm] {Autoencoder};
    \end{tikzpicture}
    \caption{Autoencoder architecture, where $\boldsymbol{u}$ is a latent variable.}
    \label{Fig:Auto}
\end{figure}

Figure \ref{Fig:Auto} shows a brief representation of an autoencoder architecture, where three states composed of functions $\boldsymbol{\upsilon}$ and $\mathcal{S}$ return an estimation of the first state, and the second state is a latent variable \cite[see][]{rodriguez2023multimodal,baig2023autoencoder}.

With all the essential components in place, the following section outlines the structure of the SITAR model integrated with an autoencoder. Our objective is to combine the flexibility and interpretability of the SITAR model with the predictive capabilities of neural networks. This integration allows us to estimate the random effects of the SITAR model and subsequently make growth curve predictions for new individuals in the population using the trained neural network.

\subsection{Deep-SITAR}\label{sec:Deep_sitar}
We extend the classical SITAR model in Equation \eqref{eq:sitar_model} to predict growth patterns for new individuals within a population by incorporating an autoencoder and unifying the mixed-effects model with the neural network framework. We refer to this modified approach as Deep-SITAR.

Let $y_{ij}$ be the observation for the $i$-th individual measured at time $t_{ij}$  , where  $1\leq i \leq N$ and  $1\leq j \leq n$. Consider the  model:
 \begin{align}\label{eq:deep_sitar_model}
     y_{ij}&=  a_i+ \mathcal{S}\left(\frac{t_{ij}- b_i}{e^{-c_i}} \right)+\epsilon_{ij},  \qquad i=1, ..., N, \quad j=1,..., n \nonumber \\
     &=  a_0+a_{1i}+\sum_{k=1}^{m}B_k\left(\frac{t_{ij}- ( b_0+b_{1i})}{e^{-( c_0+c_{1i})}} \right)\alpha_i+ \epsilon_{ij}
 \end{align}

 where  $S(\cdot)$  is a smooth and unknown function that is approximated by a cubic B-splines basis functions $\{{B}_k\}_{i=1}^m$, $\boldsymbol{\alpha}=(\alpha_1, ..., \alpha_m)^\top$ a vector of coefficients, $a_0, b_0$ and $c_0$ are fixed effects and $a_{1i}, b_{1i}$ and $c_{1i}$ are normally distributed random effects with mean $\boldsymbol{0}$ and variance-covariance matrix $\boldsymbol{\Lambda}$.  The error $\epsilon_{ij}$ is assumed to be independent of random effects and is normally distributed with mean $0$ and variance $\sigma^2$.  In this analysis, we use equally spaced knots to construct B-splines. Additionally, the size of the B-spline basis is related to the number of segments formed by the knots through the formula $m=n_{\mbox{\tiny seg}}+q$, where $n_{\mbox{\tiny seg}}$ is the number of segments, and $q$ represents the polynomial degree of the B-spline, typically set to $q=3$ \cite[see][for more details]{eilers2021practical}.

Since nonlinearity is present in Equation \eqref{eq:deep_sitar_model}, we employ neural networks to estimate the Deep-SITAR model using an autoencoder procedure.  This consists of an encoder provided by a neural network and predicts the SITAR random effects $\boldsymbol{u}_i=(a_{1i}, b_{1i},c_{1i})$, which are the autoencoder's latent variables. As a decoder, it uses a B-spline that fits individual curves $\boldsymbol{y}_i$ over time $\boldsymbol{t}_i$, given the random effects estimations obtained from the encoder. Considering the loss function for an autoencoder model shown in Equation \eqref{MSE1}, and the loss function for an neural network that adjusts random effects presented in Equation \eqref{nnre}, the proposed deep-SITAR model will have a loss function as follows:

\begin{equation}
    \mathcal{L}_{\boldsymbol{\theta}}(y,\boldsymbol{u}) := \frac{1}{N}\sum_{i=1}^N\left(\|\mathcal{S}\circ\boldsymbol{\upsilon}_{\boldsymbol{\theta}}(y_i,\boldsymbol{u}_i)-\boldsymbol{u}_i\|_2^2 +  \boldsymbol{\upsilon}_{\boldsymbol{\theta}}(\boldsymbol{y}_i)^\top\boldsymbol{\Lambda}_{\boldsymbol{ \upsilon}}^{-1}\boldsymbol{\upsilon}_{\boldsymbol{\theta}}(\boldsymbol{y}_i)\right).\label{MSEVAE}
\end{equation}

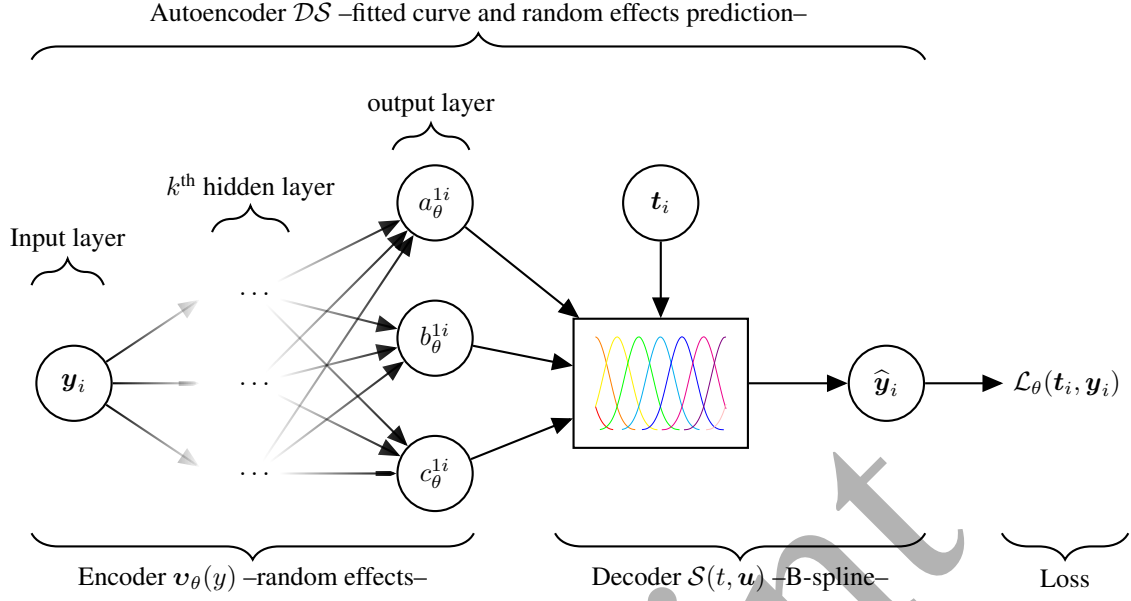
\begin{figure}[h!]
\centering
\begin{tikzpicture}[thick, scale=0.6]
    \tikzstyle{unit0}=[draw,shape=circle,minimum size=1.cm]
	\tikzstyle{unit}=[draw,shape=circle,minimum size=1.cm]
	\tikzstyle{unit2}=[draw,shape=circle,minimum size=1.3cm]
	\tikzstyle{hidden}=[draw,shape=circle,minimum size=1.cm]
	\tikzstyle{FP}=[draw,shape=rectangle,minimum size=1.3cm]

	\node[unit](x0) at (0,2){$\boldsymbol{y}_i$};
	
	\node(h22) at (3,0){};
	\node(h21) at (3,2){};
	\node(h20) at (3,4){};
	
	\node(d3) at (4,0){$\ldots$};
	\node(d2) at (4,2){$\ldots$};
	\node(d1) at (4,4){$\ldots$};

	\node(hL12) at (5,0){};
	\node(hL11) at (5,2){};
	\node(hL10) at (5,4){};
	
	\node[unit](y1) at (8,6){$a^{1i}_{\theta}$};
	\node[unit](y2) at (8,3){$b^{1i}_{\theta}$};
	\node[unit](y3) at (8,0){$c^{1i}_{\theta}$};
	
	\node[unit](FPy1) at (13,6){$\boldsymbol{t}_i$};
	\node[FP](FPy2) at (13,2){
 \begin{tikzpicture}[scale=0.2]
\begin{axis}[
    width=12cm, height=9cm, 
    axis lines=none, 
    xlabel={}, ylabel={}, 
    xtick=\empty, ytick=\empty, 
    every axis plot/.append style={thick}
]
\addplot [red, smooth] table {
0.000 0.169
0.017 0.122
0.034 0.085
0.051 0.056
0.068 0.035
0.085 0.020
0.102 0.010
0.119 0.004
0.136 0.001
};
\addplot [orange, smooth] table {
0.000 0.667
0.017 0.657
0.034 0.630
0.051 0.589
0.068 0.536
0.085 0.474
0.102 0.408
0.119 0.340
0.136 0.273
0.153 0.211
0.169 0.157
0.186 0.113
0.203 0.078
0.220 0.051
0.237 0.031
0.254 0.017
0.271 0.008
0.288 0.003
0.305 0.001
};
\addplot [yellow, smooth] table {
0.000 0.165
0.017 0.220
0.034 0.283
0.051 0.350
0.068 0.418
0.085 0.484
0.102 0.544
0.119 0.596
0.136 0.636
0.153 0.660
0.169 0.666
0.186 0.653
0.203 0.622
0.220 0.577
0.237 0.522
0.254 0.459
0.271 0.392
0.288 0.324
0.305 0.258
0.322 0.198
0.339 0.146
0.356 0.104
0.373 0.071
0.390 0.046
0.407 0.027
0.424 0.015
0.441 0.007
0.458 0.002
0.475 0.000
};
\addplot [green, smooth] table {
0.034 0.001
0.051 0.005
0.068 0.011
0.085 0.022
0.102 0.038
0.119 0.060
0.136 0.090
0.153 0.129
0.169 0.177
0.186 0.234
0.203 0.298
0.220 0.366
0.237 0.434
0.254 0.499
0.271 0.557
0.288 0.606
0.305 0.643
0.322 0.663
0.339 0.665
0.356 0.647
0.373 0.613
0.390 0.566
0.407 0.508
0.424 0.444
0.441 0.377
0.458 0.309
0.475 0.244
0.492 0.185
0.508 0.136
0.525 0.096
0.542 0.064
0.559 0.041
0.576 0.024
0.593 0.012
0.610 0.005
0.627 0.002
0.644 0.000
};
\addplot [cyan, smooth] table {
0.203 0.002
0.220 0.006
0.237 0.013
0.254 0.025
0.271 0.042
0.288 0.066
0.305 0.098
0.322 0.139
0.339 0.189
0.356 0.248
0.373 0.314
0.390 0.382
0.407 0.449
0.424 0.513
0.441 0.569
0.458 0.616
0.475 0.649
0.492 0.665
0.508 0.662
0.525 0.641
0.542 0.603
0.559 0.553
0.576 0.494
0.593 0.429
0.610 0.361
0.627 0.294
0.644 0.230
0.661 0.173
0.678 0.126
0.695 0.088
0.712 0.058
0.729 0.037
0.746 0.021
0.763 0.011
0.780 0.004
0.797 0.001
};
\addplot [blue, smooth] table {
0.356 0.001
0.373 0.003
0.390 0.007
0.407 0.015
0.424 0.028
0.441 0.047
0.458 0.073
0.475 0.107
0.492 0.149
0.508 0.202
0.525 0.263
0.542 0.329
0.559 0.397
0.576 0.464
0.593 0.526
0.610 0.581
0.627 0.625
0.644 0.654
0.661 0.667
0.678 0.659
0.695 0.633
0.712 0.593
0.729 0.540
0.746 0.480
0.763 0.414
0.780 0.345
0.797 0.279
0.814 0.216
0.831 0.161
0.847 0.116
0.864 0.080
0.881 0.053
0.898 0.032
0.915 0.018
0.932 0.009
0.949 0.003
0.966 0.001
0.983 0.000
};
\addplot [magenta, smooth] table {
0.508 0.000
0.525 0.001
0.542 0.003
0.559 0.009
0.576 0.018
0.593 0.032
0.610 0.053
0.627 0.080
0.644 0.116
0.661 0.161
0.678 0.215
0.695 0.278
0.712 0.345
0.729 0.413
0.746 0.479
0.763 0.540
0.780 0.592
0.797 0.633
0.814 0.659
0.831 0.667
0.847 0.655
0.864 0.625
0.881 0.582
0.898 0.527
0.915 0.465
0.932 0.398
0.949 0.330
0.966 0.264
0.983 0.203
1.000 0.150
};
\addplot [violet, smooth] table {
0.678 0.000
0.695 0.001
0.712 0.004
0.729 0.010
0.746 0.021
0.763 0.036
0.780 0.058
0.797 0.087
0.814 0.125
0.831 0.172
0.847 0.229
0.864 0.293
0.881 0.360
0.898 0.428
0.915 0.493
0.932 0.553
0.949 0.603
0.966 0.640
0.983 0.662
1.000 0.665
};
\addplot [pink, smooth] table {
0.847 0.000
0.864 0.002
0.881 0.005
0.898 0.012
0.915 0.024
0.932 0.041
0.949 0.064
0.966 0.095
0.983 0.135
1.000 0.185
};
\end{axis}
\end{tikzpicture}
    };
	\node[unit](ypred) at (18,2){$\widehat{\boldsymbol{y}}_i$};
 
	\node[](lo) at (22,2){$\mathcal{L}_{\theta}(\boldsymbol{t}_i,\boldsymbol{y}_i)$};
	
	\draw[->] (y1) -- (FPy2);
	\draw[->] (y2) -- (FPy2);
	\draw[->] (y3) -- (FPy2);
	
	\draw[->] (FPy1) -- (FPy2);

	\draw[->] (FPy2) -- (ypred);

        \draw[->] (ypred) -- (lo);
	
    \draw[->,path fading=east] (x0) -- (h20);
	\draw[->,path fading=east] (x0) -- (h21);
	\draw[->,path fading=east] (x0) -- (h22);
	
	\draw[->,path fading=west] (d1) -- (y1);
	\draw[->,path fading=west] (d1) -- (y2);
	\draw[->,path fading=west] (d1) -- (y3);
	
	\draw[->,path fading=west] (d2) -- (y1);
	\draw[->,path fading=west] (d2) -- (y2);
	\draw[->,path fading=west] (d2) -- (y3);
	
	\draw[->,path fading=west] (d3) -- (y1);
	\draw[->,path fading=west] (d3) -- (y2);
	\draw[->,path fading=west] (d3) -- (y3);
	
	\draw [decorate,decoration={brace,amplitude=10pt},xshift=-4pt,yshift=5pt] (-0.8,4) -- (0.8,4) node [black,midway,yshift=+0.6cm]{Input layer};
	\draw [decorate,decoration={brace,amplitude=10pt},xshift=-4pt,yshift=10pt] (3.2,5) -- (4.9,5) node [black,midway,yshift=+0.6cm]{$k^{\text{th}}$ hidden layer};
	\draw [decorate,decoration={brace,amplitude=10pt},xshift=-4pt,yshift=5pt] (7.2,7) -- (8.9,7) node [black,midway,yshift=+0.6cm]{output layer};
	\draw [decorate,decoration={brace,amplitude=10pt,mirror},xshift=-4pt,yshift=5pt] (-0.8,-1.5) -- (9,-1.5) node [black,midway,yshift=-0.6cm]{Encoder $\boldsymbol{\upsilon}_{\theta}(y)$ --random effects--};
	\draw [decorate,decoration={brace,amplitude=10pt,mirror},xshift=-4pt,yshift=5pt] (10.8,-1.5) -- (19,-1.5) node [black,midway,yshift=-0.6cm]{Decoder $\mathcal{S}(t,\boldsymbol{u})$ --B-spline--};
	\draw [decorate,decoration={brace,amplitude=10pt,mirror},xshift=-4pt,yshift=5pt] (20.7,-1.5) -- (23.5,-1.5) node [black,midway,yshift=-0.6cm]{Loss};
    \draw [decorate,decoration={brace,amplitude=10pt},xshift=-4pt,yshift=5pt] (-.8,9) -- (19,9) node [black,midway,yshift=+0.6cm]{Autoencoder  $\mathcal{DS}$ --fitted curve and random effects prediction--};
\end{tikzpicture}
\caption{Deep-SITAR scheme where the output is an input estimation, and ${\boldsymbol{u}}_i=(a_{\theta}^{1i},b_{\theta}^{1i},c_{\theta}^{1i})$.}
\label{fig:E-D1}
\end{figure}

Figure \ref{fig:E-D1} illustrates this process, and Algorithm \ref{alg:deep_sitar} outlines the steps for estimating Deep-SITAR parameters. Estimation is performed using Stochastic Gradient Descent (SGD) to minimize Equation \eqref{MSEVAE}.

\RestyleAlgo{ruled}
\SetKwComment{Comment}{/* }{ */}
\begin{algorithm}[H]
\caption{Deep-SITAR with SGD}\label{alg:deep_sitar}
\KwData{$t$, $\{\boldsymbol{y}_i\}_{i=1}^N$, $n_{\mbox{\tiny seg}}, \boldsymbol{\theta}_0$, $epochs$ \Comment*[r]{Initial value}
    }
\KwResult{$\{\widehat{\boldsymbol{y}}_i\}_{i=1}^N$, $\widehat{\theta}$, $\{a^i_{\widehat{\theta}}\}_{i=1}^N$, $\{b^i_{\widehat{\theta}}\}_{i=1}^N$, $\{c^i_{\widehat{\theta}}\}_{i=1}^N$  \Comment*[r]{Estimation}
}
$k \gets 0$\;
\For{$k \gets 1$ \textbf{to} epochs}{
\For{$i \gets 1$ \textbf{to} $N$}{
$a_k^i \gets a_{\theta_k}^i$\;
$b_k^i \gets b_{\theta_k}^i$\;
$c_k^i \gets c_{\theta_k}^i$\;
$\boldsymbol{y}^k_i \gets a_k^i+\mathcal{S}\left(\frac{\boldsymbol{t}_i + b_k^i}{\exp{(c_k^i)}};n_{\mbox{\tiny seg}}\right)$\;
}
$\Sigma \gets Cov\left(\{a_k^i\}_{i=1}^N,\{b_k^i\}_{i=1}^N,\{c_k^i\}_{i=1}^N\right)$\;
$\nabla \mathcal{L} \gets 0$\;
\For{$i \gets 1$ \textbf{to} $N$}{
$\delta_i \gets [a_k^i, b_k^i, c_k^i]\boldsymbol{\Lambda}^{-1}[a_k^i, b_k^i, c_k^i]^T$\;
$\nabla \mathcal{L} \gets \nabla \mathcal{L} + \frac{\partial}{\partial \theta}\left(\|\boldsymbol{y}_i-\boldsymbol{y}_i^k\|^2_2 + \delta_i\right)$\;
}
$\boldsymbol{\theta}_{k+1} \gets \boldsymbol{\theta}_k - \alpha\nabla \mathcal{L}$\;
$k \gets k + 1$\;
}
$\widehat{\boldsymbol{\theta}} \gets \boldsymbol{\theta}_{epochs}$\;
$\{a^i_{\widehat{\theta}}\}_{i=1}^N \gets \{a_{\theta_{epochs}}^i\}_{i=1}^N$\;
$\{b^i_{\widehat{\theta}}\}_{i=1}^N \gets \{b_{\theta_{epochs}}^i\}_{i=1}^N$\;
$\{c^i_{\widehat{\theta}}\}_{i=1}^N \gets \{c_{\theta_{epochs}}^i\}_{i=1}^N$\;
$\{\widehat{\boldsymbol{y}}_i\}_{i=1}^N \gets \left\{a^i_{\widehat{\theta}}+\mathcal{S}\left(\frac{\boldsymbol{t}_i + b^i_{\widehat{\theta}}}{\exp{(c^i_{\widehat{\theta}})}};n_{\mbox{\tiny seg}} \right)\right\}_{i=1}^N$\;
\textbf{return }$\{\widehat{\boldsymbol{y}}_i\}_{i=1}^N$, $\widehat{\theta}$, $\{a^i_{\widehat{\theta}}\}_{i=1}^N$, $\{b^i_{\widehat{\theta}}\}_{i=1}^N$, $\{c^i_{\widehat{\theta}}\}_{i=1}^N$
\end{algorithm}

\medskip

We have introduced the Deep-SITAR model and provided its implementation. In the following section, we present an experiment using simulated data to evaluate the performance of Deep-SITAR in predicting growth curves. Additionally, we compare its curve-fitting capabilities with those of SITAR.

\section{Experiment}\label{sec:simulation}

\subsection{Data and model description}\label{sec:data}

The experiment is designed to assess the performance of the proposed Deep-SITAR and compare it with the classical SITAR. To achieve this, three data sets were generated, varying the number of individuals $(N = 500, 1000, 5000)$. Each data set is divided into 80\% for training and 20\% for validation. The heights are generated based on the growth curves of the Berkeley data set, available in the \texttt{sitar} library of \texttt{R}, following these steps:
\begin{enumerate}
    \item  Twenty ages, equally spaced between 9 and 18 years, are generated for each individual while maintaining a balanced design. 
    \item  SITAR is fitted to the Berkeley dataset with $df = 5$ to obtain estimates of $a_0,\ b_0,\ c_0$ and $\boldsymbol{\Lambda}$.
    \item Using the estimates obtained in Step 2 and adding a noise term $\epsilon_{ij} \sim \mathcal{N}(0, 0.4)$  the heights are generated according to:
 $$   y_{ij}= a_i + \sum_{k=1}^5 B_k\left(\frac{t_{ij}- b_i}{e^{-c_i}}\right)+\epsilon_{ij} \quad  ,1\leq i \leq N, \ \ 1 \leq j \leq 20 $$

 where $a_i, \ b_i $ and $c_i$ are the individual parameters for the $i$-th individual and $\{B_k\}^5_{k=1}$ is a natural cubic B-spline basis functions. Note that the estimates obtained in Step 2 serve as the true parameter values for model evaluation.
\end{enumerate}

To implement the Deep-SITAR model, as illustrated in Figure \ref{fig:E-D1}, we use a cubic B-spline basis function, varying the number of segments $n_{\mbox{\tiny seg}}= 5,8,10,15$. Moreover, Table \ref{tab:neural network} presents the neural network architecture used in the simulation.  For the SITAR model, $df$ is selected using the BIC criterion in each sample $N$.

\begin{table}[H]
\centering
\begin{tabular}{|l|l|l|}
\hline
\multicolumn{3}{|c|}{\textbf{Deep-SITAR encoder}} \\ \hline
\textbf{Layer (type)} & \textbf{Output Shape} & \textbf{Activation Function} \\
\hline \hline
Input non-trainable layer ($\boldsymbol{y}_i$) & (None, $20$) & \emph{linear} \\ \hline
Hidden Dense layer & (None, 30) & \emph{tanh} \\ \hline
Hidden Dense layer & (None, 30) & \emph{tanh} \\ \hline
Output Dense layer ($a_{1i},b_{1i},c_{1i}$) & (None, $3$) & \emph{linear} \\ 
\hline
\end{tabular}
\caption{Architecture for Deep-SITAR encoder, where the input is the height of each individual ($\boldsymbol{y}_i$), and output is the random effects $\boldsymbol{u}_i=(a_{1i},b_{1i},c_{1i})$.}
\label{tab:neural network}
\end{table}

\subsection{Results}

We fitted the SITAR and Deep-SITAR models to the simulated growth curves described in section \ref{sec:data}. The $df$ parameter value for fitting SITAR is selected using the Bayesian Information Criterion (BIC) over a range of $df$ from 3 to 15. For sample sizes $N$ of $500, 1000,$ and $5000$ individuals, the selected models had $df = 7$, $df = 5$, and $df = 5$, respectively. For Deep-SITAR, we use the neural network architecture in Table \ref{tab:neural network} and tested four different values for the number of segments for the B-spline basis: $n_{\mbox{\tiny seg}} = 5, 8, 10$, and $15$.

\begin{figure}[!htb]
\centering
\includegraphics[width=1\textwidth, page=1]{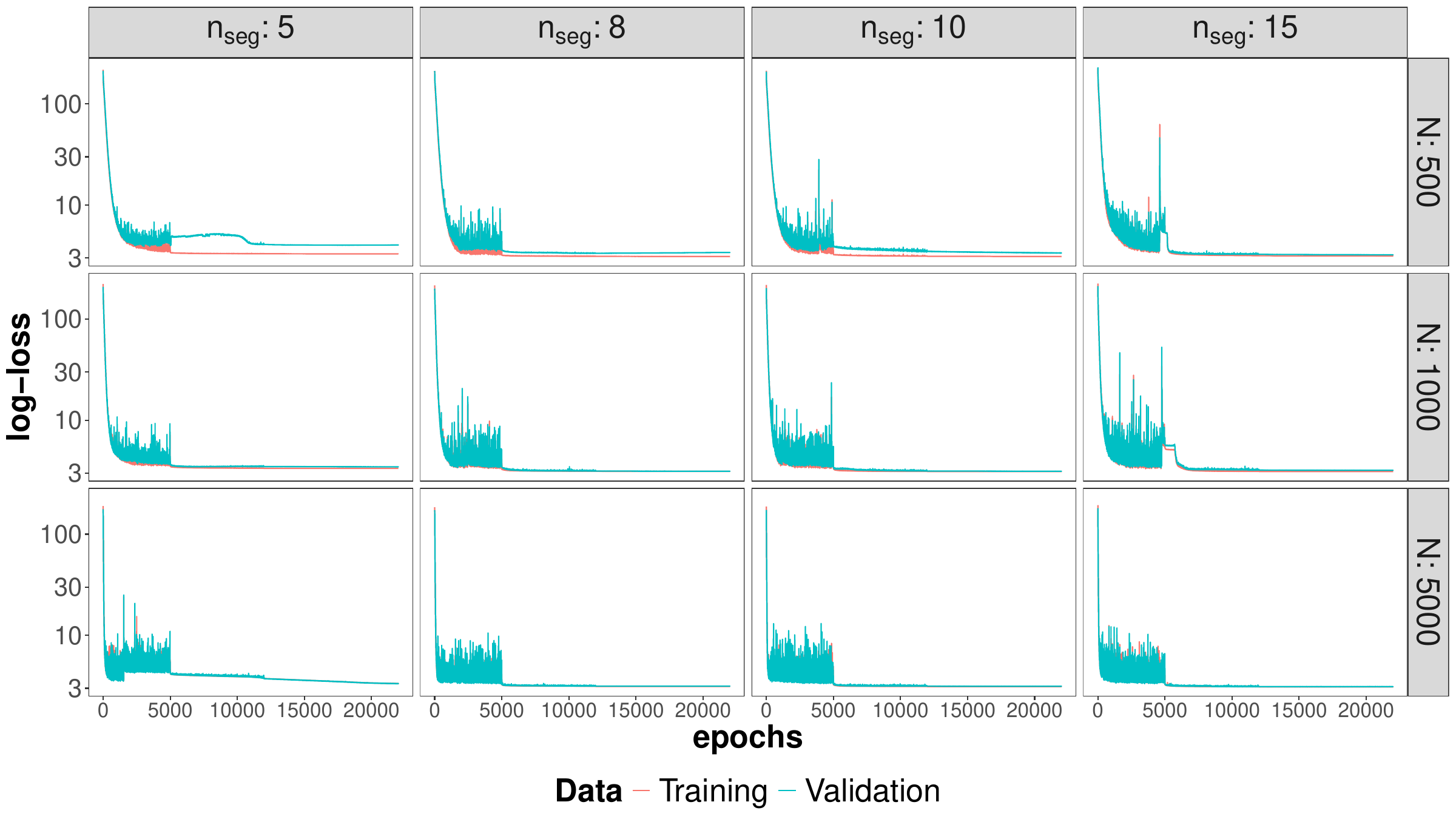} 
\caption{Log-loss for the Deep-SITAR approach for the training and validation datasets, varying by sample size $N$ and number of segments $n_{\mbox{\tiny seg}}$ in the B-spline}\label{fig:loss_all}
\end{figure}

Figure \ref{fig:loss_all} shows the log-loss behavior for the training and validation datasets obtained for Deep-SITAR. The log-loss begins to stabilize around iteration 1400, with no signs of overfitting, as the gap between training and validation loss remains small. Table \ref{tab:loss} presents the log-loss values for training and validation datasets at the last iteration. Increasing $n_{\mbox{\tiny seg}}$ generally leads to lower log-loss, and as neural networks grow, the difference between training and validation log-loss narrows. This pattern reinforces the idea that a more extensive dataset improves generalization and enhances predictive performance on new data.

\begin{figure}[!htb]
\centering
\includegraphics[width=1\textwidth, page=1]{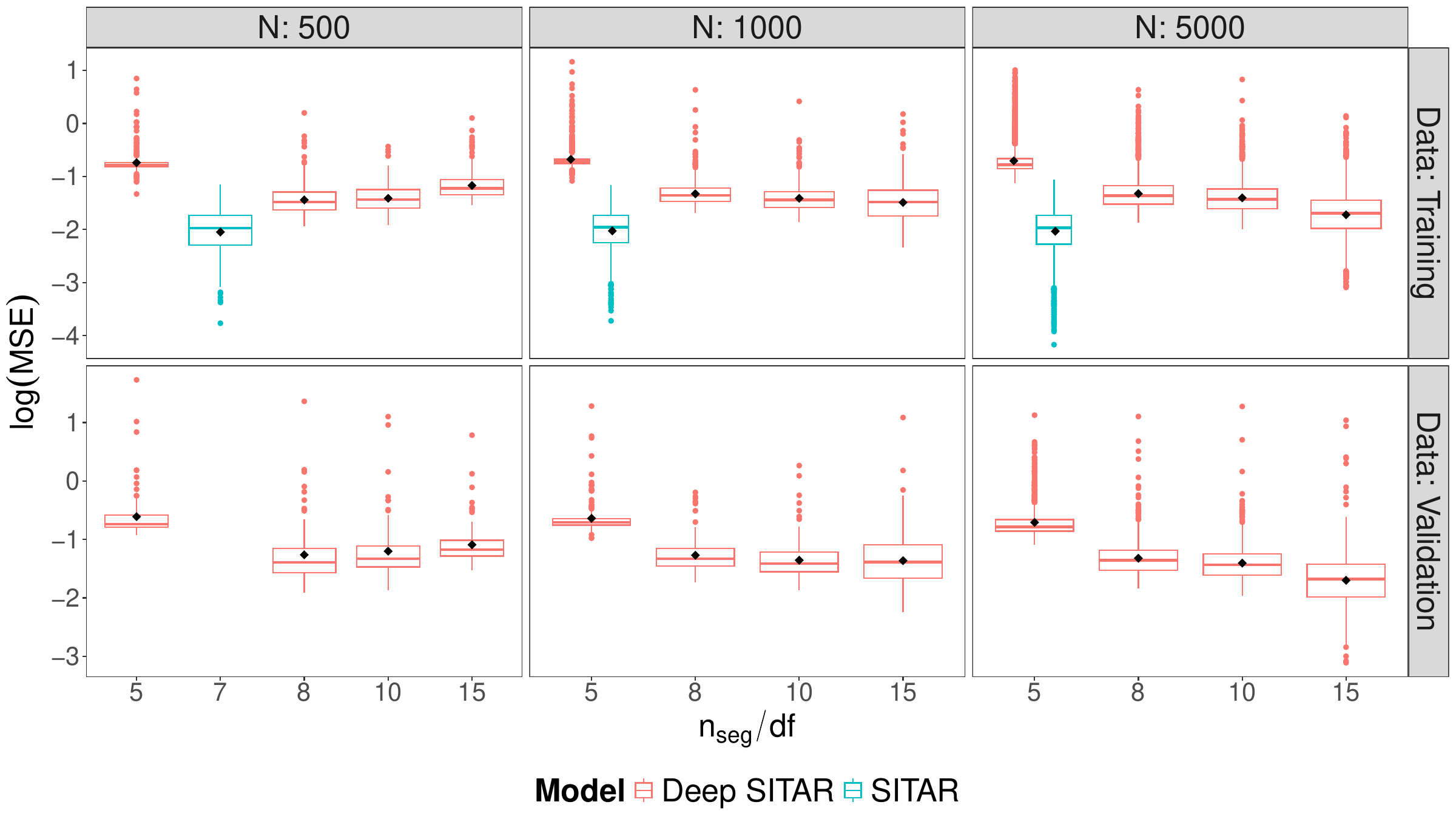} 
\caption{$\log(\mbox{MSE})$ individual curves for SITAR and Deep-SITAR approaches in the sample $N$ by a number of segments $n_{\mbox{\tiny \emph{seg}}}$ for the B-spline. The black point represents the average. }
\label{fig:log_mse}
\end{figure}

The results assessing the performance of Deep-SITAR and comparing its curve-fitting capabilities with SITAR are presented in Table \ref{tab:mae} and Figure \ref{fig:log_mse}. These results show that increasing the sample size $N$ and number of segments $n_{\mbox{\tiny seg}}$ for Deep-SITAR improves model performance on average, with the best results achieved at $N=5000$ and $n_{\mbox{\tiny seg}}=15$. In addition, Deep-SITAR displays better generalization with a larger sample size and more segments for the B-spline. However, MSE in the validation set is quite variable, particularly when $N=500$. SITAR consistently achieves the lowest training mean MSE with values ranging from 0.013 to 0.014, significantly lower than Deep-SITAR's. However, the SITAR does not provide validation results and predictions for new individuals.

\begin{figure}[!htb]
\centering
\includegraphics[width=1\textwidth]{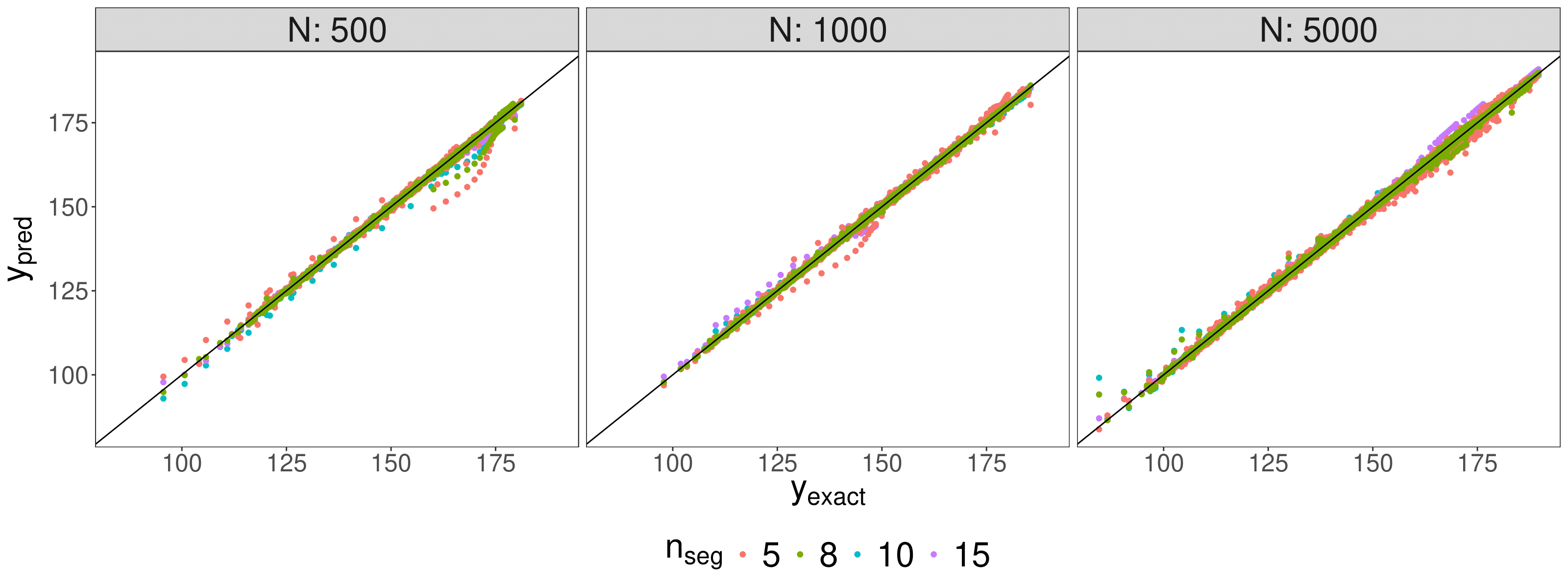} 
\caption{ Comparison of predicted values $y_{\mbox{\tiny \emph{pred}}}$ versus exact values without noise $y_{\mbox{\tiny \emph{exact}}}$ for the Deep-SITAR in the validation dataset.}
\label{fig:pred_eval}
\end{figure}

Figure \ref{fig:pred_eval} compares the fit of Deep-SITAR with the exact values --without noise-- from the validation set, varying the number of $n_{\mbox{\tiny seg}}$. We observe that the fit improves as the number of segments increases. However, fitting issues tend to arise at the boundaries of the curves, especially with smaller sample sizes and fewer segments. Regarding the variance estimation of the random effect, Table \ref{tab:var} shows that as the sample size $N$ increases, the absolute differences for all parameters tend to decrease for both Deep-SITAR and SITAR. However, more significant absolute errors persist in estimating $\sigma_a^2$, particularly for smaller sample sizes. SITAR shows consistently low differences for all random effects variances across all sample sizes, typically close to $0$ for $\sigma_b^2$ and $\sigma_c^2$. Meanwhile, Deep-SITAR performs well in estimating random effect variances, with the most accurate estimations occurring at more extensive sample sizes. 

\begin{table}[!h]
\centering
\resizebox{0.45\textwidth}{!}{
\begin{tabular}[t]{cccc}
\toprule
N & n$_{\mbox{\tiny seg}}$ & log-loss$_{\mbox{\tiny \emph{Training}}}$ & log-loss$_{\mbox{\tiny \emph{Validation}}}$ \\
\midrule
 & 5 & 3.3004 & 4.0377\\

 & 8 & 3.1132 & 3.4030\\

 & 10 & 3.1105 & 3.3672\\

\multirow{-4}{*}{\centering\arraybackslash 500} & 15 & 3.1465 & 3.2289\\
\cmidrule{1-4}
 & 5 & 3.3434 & 3.4688\\

 & 8 & 3.1266 & 3.1348\\

 & 10 & 3.1132 & 3.1359\\

\multirow{-4}{*}{\centering\arraybackslash 1000} & 15 & 3.1130 & 3.2015\\
\cmidrule{1-4}
 & 5 & 3.3450 & 3.3351\\

 & 8 & 3.1293 & 3.1492\\

 & 10 & 3.1166 & 3.1415\\

\multirow{-4}{*}{\centering\arraybackslash 5000} & 15 & 3.0946 & 3.1228\\
\bottomrule
\end{tabular}}
\caption{Results of log-loss for the last iteration $It=22000$ by sample size $(N)$ and number of segments $(n_{\mbox{\tiny seg}})$ for the B-spline basis  }
\label{tab:loss}
\end{table}


\begin{table}[!h]
\centering
\resizebox{0.65\textwidth}{!}{
\begin{tabular}[t]{ccccccc}
\toprule
\multicolumn{3}{c}{\textbf{ }} & \multicolumn{2}{c}{\textbf{ Training}} & \multicolumn{2}{c}{\textbf{Validation}} \\
\cmidrule(l{3pt}r{3pt}){4-5} \cmidrule(l{3pt}r{3pt}){6-7}
N & Model & n$_{\mbox{\tiny \emph{seg}}}$ & mean$_{\mbox{\tiny \emph{MSE}}}$ & sd$_{\mbox{\tiny \emph{MSE}}}$ & mean$_{\mbox{\tiny \emph{MSE}}}$ & sd$_{\mbox{\tiny \emph{MSE}}}$\\
\midrule
 &  & 5 & 0.233 & 0.463 & 0.963 & 5.473\\

 &  & 8 & 0.050 & 0.092 & 0.336 & 2.309\\

 &  & 10 & 0.047 & 0.038 & 0.310 & 1.546\\

 & \multirow{-4}{*}{\centering\arraybackslash Deep-SITAR} & 15 & 0.084 & 0.093 & 0.167 & 0.619\\
\cmidrule{2-7}
\multirow{-5}{*}{\centering\arraybackslash 500} & SITAR & 7 & 0.013 & 0.011 & - & -\\
\cmidrule{1-7}
 &  & 5 & 0.279 & 0.689 & 0.397 & 1.456\\

 &  & 8 & 0.062 & 0.170 & 0.070 & 0.084\\

 &  & 10 & 0.049 & 0.099 & 0.069 & 0.162\\

 & \multirow{-4}{*}{\centering\arraybackslash Deep-SITAR} & 15 & 0.049 & 0.083 & 0.133 & 0.868\\
\cmidrule{2-7}
\multirow{-5}{*}{\centering\arraybackslash 1000} & SITAR & 5 & 0.014 & 0.011 & - & -\\
\cmidrule{1-7}
 &  & 5 & 0.279 & 0.566 & 0.274 & 0.579\\

 &  & 8 & 0.066 & 0.130 & 0.083 & 0.450\\

 &  & 10 & 0.053 & 0.128 & 0.075 & 0.619\\

 & \multirow{-4}{*}{\centering\arraybackslash Deep-SITAR} & 15 & 0.031 & 0.055 & 0.057 & 0.459\\
\cmidrule{2-7}
\multirow{-5}{*}{\centering\arraybackslash 5000} & SITAR & 5 & 0.014 & 0.011 & - & -\\
\bottomrule
\end{tabular}}
\caption{ Results of the average MSE of individual curve estimations for SITAR and Deep-SITAR approaches by sample size $(N)$, Model and number of segments $(n_{\mbox{\tiny seg}})$ for the B-spline basis.}
\label{tab:mae}
\end{table}
\begin{table}[!h]
\centering
\resizebox{0.6\textwidth}{!}{
\begin{tabular}[t]{ccccccccc}
\toprule
N & Model & $n_{\mbox{\tiny \emph{seg}}}$ & $|\sigma_a^2-\hat{\sigma}_a^2|$ & $|\sigma_b^2-\hat{\sigma}_b^2|$ & $|\sigma_c^2-\hat{\sigma}_c^2|$\\
\midrule
 &  & 5 & 4.423 & 1.016 & 0.012\\

 &  & 8 & 4.626 & 0.186 & 0.001\\

 &  & 10 & 4.696 & 0.502 & 0.001\\

 & \multirow{-4}{*}{\centering\arraybackslash Deep-SITAR} & 15 & 4.630 & 1.447 & 0.004\\
\cmidrule{2-6}
\multirow{-5}{*}{\centering\arraybackslash 500} & SITAR & 7 & 4.660 & 0.079 & 0.001\\
\cmidrule{1-6}
 &  & 5 & 2.235 & 1.080 & 0.018\\

 &  & 8 & 2.701 & 0.016 & 0.007\\

 &  & 10 & 2.675 & 0.552 & 0.007\\

 & \multirow{-4}{*}{\centering\arraybackslash Deep-SITAR} & 15 & 2.836 & 1.893 & 0.006\\
\cmidrule{2-6}
\multirow{-5}{*}{\centering\arraybackslash 1000} & SITAR & 5 & 2.601 & 0.062 & 0.005\\
\cmidrule{1-6}
 &  & 5 & 1.197 & 0.896 & 0.013\\

 &  & 8 & 1.052 & 0.104 & 0.002\\

 &  & 10 & 0.811 & 0.529 & 0.000\\

 & \multirow{-4}{*}{\centering\arraybackslash Deep-SITAR} & 15 & 1.132 & 0.133 & 0.001\\
\cmidrule{2-6}
\multirow{-5}{*}{\centering\arraybackslash 5000} & SITAR & 5 & 0.888 & 0.002 & 0.002\\
\bottomrule
\end{tabular}}
\caption{Results of absolute differences between the estimates and true values of the random effect variance for SITAR and Deep-SITAR, grouped by simple size $N$ and the number of segments $n_{\mbox{\tiny \emph{seg}}}$ for the B-spline basis.}
\label{tab:var}
\end{table}

\section{Conclusion } \label{sec:conclusion}

The proposed Deep-SITAR model extends the classical SITAR by integrating it into a neural network framework via autoencoders. The model is designed to combine the flexibility and interpretability of the SITAR model with the predictive power of neural networks, using an autoencoder to integrate the neural network with the B-spline and simulate the classical SITAR. The proposed neural network to estimate the random effects provides us with a novel form to obtain it, changing the classical estimation by an SGD and obtaining a model for future individual predictions.

\clearpage
\newpage
In our experiment, as expected, the SITAR model provided consistently stable and reliable results across different sample sizes. In contrast, the Deep-SITAR model, which includes a neural network component --usually loses precision to gain predictions and robustness-- exhibited more variability in the performance of MSE than SITAR. However, Deep-SITAR demonstrated considerable improvements in the loss curve and the MAE with larger sample sizes $N$ and an increasing number of segments $n_{\mbox{\tiny seg}}$ for the B-spline basis. Moreover, it showed a good fitting of individual curves in the different samples for the training data and a remarkable ability to make accurate predictions on the validation set, providing reliable growth curve predictions for new individuals.

The Deep-SITAR model's performance tends to improve with larger datasets, making it a strong candidate for situations where large datasets are available and the data complexity requires a more flexible model. For future work, it is important to investigate the Deep-SITAR model's behavior in estimating the derivatives of the growth curves, which are crucial for growth analysis.

\section{Funding}
This research was supported in part by the project PID2020-115882RB-I00 and 
by the Spanish Ministry of Science, Innovation, and Universities through BCAM Severo Ochoa
accreditation SEV-2017-0718 and PRE2019-089900 funding.

\section{Data availability}

The Deep-SITAR code used in this work is available on GitHub repository \url{https://github.com/orodriguezm1/DeepSITAR}.


\bibliographystyle{plain}

\end{document}